**Title**
Geospatial Knowledge Graphs

**Your Name** Rui Zhu
**Affiliation** University of Bristol
**Email Address** rui.zhu@bristol.ac.uk

**Word Count**
2749

**Abstract**
Geospatial knowledge graphs have emerged as a novel paradigm for representing and reasoning over geospatial information. In this framework, entities such as places, people, events, and observations are depicted as nodes, while their relationships are represented as edges. This graph-based data format lays the foundation for creating a "FAIR" (Findable, Accessible, Interoperable, and Reusable) environment, facilitating the management and analysis of geographic information. This entry first introduces key concepts in knowledge graphs along with their associated standardization and tools. It then delves into the application of knowledge graphs in geography and environmental sciences, emphasizing their role in bridging symbolic and subsymbolic GeoAI to address cross-disciplinary geospatial challenges. At the end, new research directions related to geospatial knowledge graphs are outlined.

**Main Text**
Knowledge graphs, and their associated technologies, provide a new way of organizing and analyzing data. In contrast to relational databases, where data are organized as tables with rows of data records and columns of associated attributes, knowledge graphs emphasize the relationships between data, with data records being represented as nodes and their relationships as edges. More formally, such a node-edge format, <node1, edge, node2> or <subject, predicate, object>, comprises the fundamental data structure of a knowledge graph, called a *triple*. Figure 1 illustrates the difference of representing the same data set in two unique ways – using a table and a graph. The data is a small sample of storms that had impacted different regions of the UK in 2023. Even though both methods can represent the same data set, a graph format outperforms in terms of their *expressiveness*. First, using a knowledge graph helps one understand the (spatial) relationships between the impacted regions, e.g., South-west Scotland is part of Scotland, which is further part of the United Kingdom together with Wales, England, and Northern Ireland. Secondly, it helps link data records from various angles. For instance, dense links between regions and storms in Figure 1 enable one to readily query all the storms that have impacted a specific region over time, with which deeper insights might be uncovered (e.g. causal relations between these hazard events). Thirdly, the ontology underlying such a graph facilitates users, especially those without much domain knowledge, to understand the meaning of a storm, the property of maximum wind gust and the definition of its units, and so on. For example, the relation of **MaxGuest** can be formalized by defining its domain as the class of **Storm**, and its range to be a data type value with a unit of knot, which is further defined by a standard, e.g. QUDT (Quantities,

Units, Dimensions, and Types Ontology), that can be linked to the graph too. Such rich expressiveness, provided by dense links and standardization of semantics, outperforms the conventional tabular view of data, empowering individuals to uncover hidden insights that might remain undiscovered otherwise.

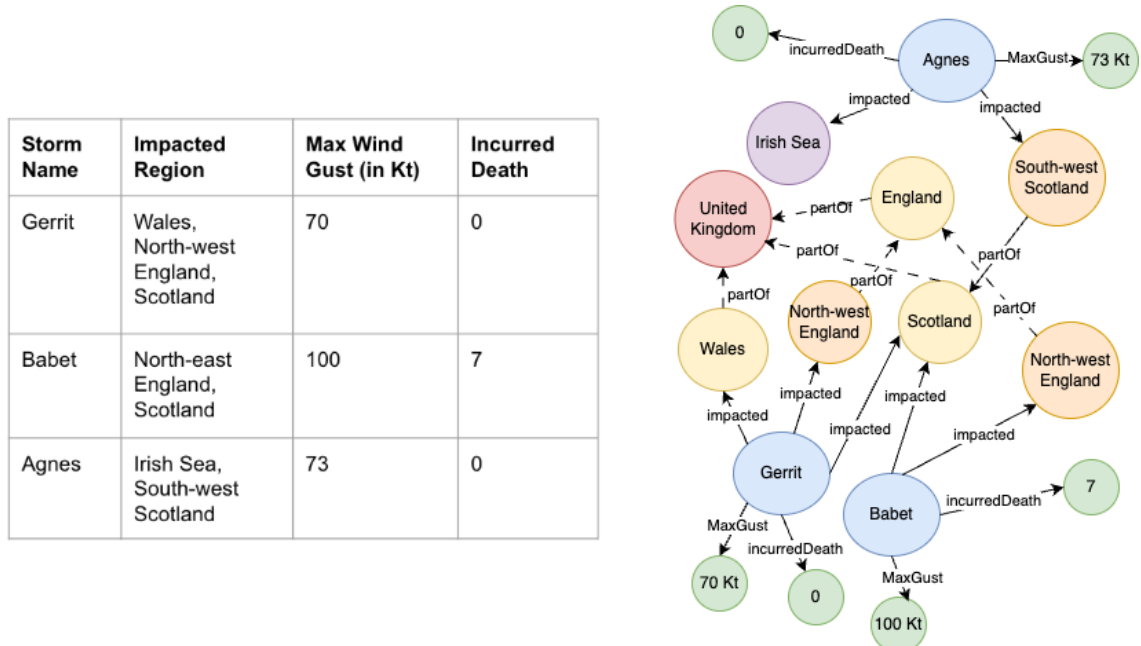

Figure 1. Tabular view (left) and graph view (right) of sampled storms affecting the UK in 2023. In the graph view, the dashed arrows indicate relations not listed in the tabular view. Note that the underlying data models have been simplified for illustration purposes.

Furthermore, knowledge graphs can be readily extended by aligning or creating new relationships between nodes. To take Figure 1 as an example, the graph can be extended by aligning the node of **Gerrit** with its correspondence in another graph, which might list its trajectory over time for instance. Also, the node of **Wales** can be linked with other datasets, such as the census data, by creating a new *hasPopulation* relation. By doing so, knowledge in the graph keeps growing, and can be processed in a homogenous environment across the boundary of different domains. Finally, as data in a knowledge graph is structured as individual statements (i.e., triples) and contextualized with rich links and standards, it becomes both human and machine-readable. This inherent AI readiness allows for direct answers to questions that traditional databases would find challenging to address. Notably, many cutting-edge information retrieval and question-answering systems, including Google's Search Engine and Apple's Siri, rely on knowledge graphs as their foundational database.

There are different techniques to implement knowledge graphs. Currently, the most notable ones are RDF graph and property graph. RDF (Resource Description Framework) was invented as a W3C (World Wide Web Consortium) standard to build an interoperable Semantic Web where contents in web pages are standardized and connected. In an RDF graph, a statement is comprised of a *subject* node, a *predicate* that describes the relationship, and an *object* node. Both nodes and edges are uniquely and identically referenced using a URI (Uniform Resource Identifier), while the object node can also be a numeric or literal value. Coupled with RDF, ontologies define the underlying schema of a graph and the most widely used standards include OWL (Web Ontology Language) and RDFS (RDF Schema). RDF graphs can be serialized into various data formats, such as Turtle, N-triples, and JSON-LD. Most graph databases (or called triple stores), including GraphDB, Stardog, and BlazeGraph, support inputting and outputting RDF graphs into data of these formats. Like using SQL to access relational databases, SPARQL is defined by W3C to query data from RDF graphs. For the validation of RDF graphs, W3C recommends using SHACL (Shape Constraint Language), enabling users to define constraints that a dataset must adhere to, with the capability to report any violations. It is noteworthy that while RDFS, OWL, and SHACL all contribute to defining the structure of an RDF graph, the former two primarily focus on reasoning and inferencing, while the latter is dedicated to data validation.

Similar to RDF graph, a property graph represents different types of entities as nodes and their connections as edges. In essence, both data models can be regarded as directed labeled graph. In contrast to RDF graph, where no properties are associated with edges, property graphs facilitate both nodes and edges with a collection of properties, which are represented as key-value pairs. This distinction allows a property graph to accommodate a flexible internal structure for both nodes and edges. However, property graphs do not possess the capability to formally model semantics, a crucial aspect for achieving data interoperability and reasoning. Instead, property graphs are designed for faster queries, utilizing Cypher as the querying language. This entry uses the term "knowledge graph" to denote an RDF graph.

Knowledge graphs have been extensively used in geography and environmental sciences for the past few years to address global interdisciplinary challenges by improving data interoperability. Hence, geospatial knowledge graphs have currently emerged as an active research field in geography, and particularly in GIScience and GeoAI. Broadly speaking, geospatial knowledge graphs refer to those specialized graphs whose nodes and/or edges are geospatial or can be georererenced. They organize statements that are about people, place, event, time, environment, scientific observation, and sensor, to name a few, and their (spatial and temporal) relationships, such as distance, direction, and topological relation. Indeed, finding a knowledge graph that is entirely non-spatial is challenging, since everything happens at some places during some time period. Furthermore, prioritizing relationships over attributes in knowledge graphs aligns with the nature of geography that "everything is related to everything else". Finally, this novel data model presents an opportunity to tackle the challenge faced by most GIS systems, which struggle to represent both quantitative and qualitative geographic information. Therefore, it is unsurprising that geographers are at the forefront of advancing this graph technology.

Knowledge representation and reasoning are two core research areas in geospatial knowledge graphs. Knowledge representation refers to the design of ontologies (or schema) for structuring data in a knowledge graph. In fact, long before geospatial knowledge graphs became the mainstream of research, geospatial ontology, or geospatial semantics more broadly, has been intensively studied setting the foundation of GIScience since 1990s (Mark et al., 1999, Mark et al., 2004, Couclelis, 2010), and the idea of building a Semantic Geospatial Web was proposed alongside the invention of the Semantic Web (Egenhofer, 2002). Recently, GeoSPARQL and OWL-Time are the two widely applied standards recommended by W3C and OGC (Open Geospatial Consortium) to represent geospatial and temporal information in knowledge graphs, respectively. Figure 2 illustrates classes defined in GeoSPARQL and they can be used to formally define most types of geographic information. For instance, a building can be defined as an instance of class **Feature**, and it has a geometry (linked using the ***hasGeometry*** predicate) that is represented as a Well-Known Text (WKT). In addition to modeling spatial objects that have associated geometry, GeoSPARQL also supports expressing spatial objects that have no geometry by simply using the **Feature** class without relating it to **Geometry**. Such a design is particularly useful for modeling human geography concepts, such as *places* which, by definition, might not necessarily be associated with a specific location with crisp boundary.

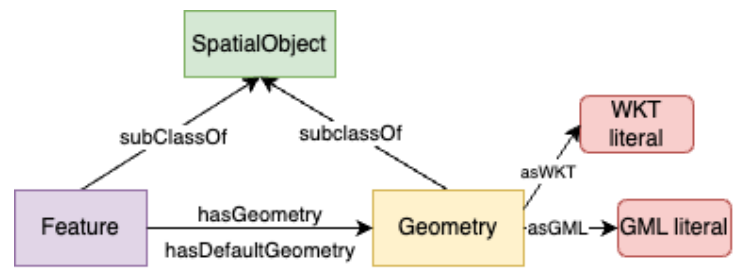

Figure 2. Diagram of GeoSPARQL (simplified version).

Leveraging theoretical foundations from geospatial semantics and advanced techniques in knowledge graphs, a growing body of research is dedicated to constructing geospatial knowledge graphs that integrate datasets across various domains, including urban studies, climate science, disaster management, agriculture, humanities, and more. One example is the project of KnowWhereGraph, utilizing space and time as the nexus to bridge diverse, multimodal, and cross-domain data "silos," thereby facilitating environmental intelligence (Janowicz et al., 2022). Its key contribution lies in the introduction of a new geospatial ontology, built on top of GeoSPARQL, to seamlessly integrate various of regional and platial identifiers, such as administrative regions, climate zones, remotely sensed image cells, and so on. With its extensive spatial coverage and comprehensive semantic modeling, KnowWhereGraph offers situational awareness for decision-makers regarding any specific location of interest. It further facilitates downstream projects in constructing their own geospatial knowledge graph by extracting from or extending the KnowWhereGraph. The Urban Flooding Open Knowledge Network (UFOKN) serves as another example, specifically concentrating on the domain of floods (Johnson et al., 2022). UFOKN combines data describing different aspects of cities with hydrologic modeling using knowledge graph, aiming to deliver real-time flood forecasting across different spatial scales. Moreover, gazetteers like GeoNames and the Geographic Names Information System (GNIS) also present their data in the knowledge graph format to cater to the community's increasing demand for interlinking data through geography.

In addition to constructing new geospatial knowledge graphs, researchers in geography also find value in open-source knowledge graphs not explicitly designed for geospatial information but containing rich contextual data for geospatial modeling. Examples include DBpedia, Wikidata, and Yago. The former two are primarily structured from Wikipedia, an open encyclopedia contributed to by volunteers globally, and Yago further structures Wikidata using schema.org, a standard ontology widely used in industry. These generalized knowledge graphs encompass geospatial entities such as cities, people, and events, often serving as the linking nexus between different knowledge graphs. For example, KnowWhereGraph is linked to Wikidata through their common entities of administrative regions, enabling the two databases to complement each other in building a comprehensive knowledge base for domain scientists.

Like geospatial representation, the exploration of reasoning over spatial and temporal information has also been a vibrant area of research in GIScience. Seminal works, such as RCC8 (Region Connection Calculus) and DE-9IM (Dimensionally Extended 9-Intersection Model), have laid the groundwork for qualitative spatial reasoning, a technique extensively applied not only in geography but also in cognitive science and artificial intelligence (AI). Specifically within knowledge graphs, RCC8 is incorporated into GeoSPARQL to standardize the querying of geospatial information using qualitative spatial relations.

These models, however, are primarily grounded in symbolic AI, where knowledge is structured using high-level symbolic representation and logic. The emergence of geospatial knowledge graphs introduces a novel approach to combine these symbolic methods with subsymbolic, or connectionist, approaches. For instance, Cai et al. (2022) investigated the potential of machine learning to learn symbolic rules for both spatial and temporal reasoning by utilizing knowledge graph embedding techniques. Additionally, Zhu et al. (2022) devised a neural network architecture for reasoning over higher-order spatial relations represented in a geospatial knowledge graph. These works combine theories from qualitative spatial reasoning with cutting-edge deep learning methods through the framework of knowledge graphs, and they pave the way towards building a neurosymbolic GeoAI where both theory and data can be integrated to responsibly guide artificial intelligent systems in processing geospatial data.

Knowledge graph embedding (KGE) is among the most popular subsymbolic methods over knowledge graphs. It is principally used to transfer the symbolic representation of nodes and edges into a continuous vector space by preserving their inherent structure and semantics in the graph. Since numeric vectors are easier to manipulate, such as computing similarity, KGE enables a wide range of applications using knowledge graphs, including link prediction, entity classification, and question answering. Regardless of the diverse downstream applications, most KGE methods can be grouped into either translation distance models (e.g., TransE) or semantic matching models (e.g., RESCAL). The key difference between these two is their scoring functions. Specifically, TransE represents both nodes as vectors: $\vec{s}$ and $\vec{o}$, and it attempts to minimize the translation error from $\vec{s}$ to $\vec{o}$ through the relation vector $\vec{r}$ (i.e., to minimize $\|\vec{s} + \vec{r} - \vec{o}\|$), if <s, r, o> holds true in the graph. In contrast, RESCAL represents the relation r as a matrix $M_r$, and the goal is to match the semantics of s and o, each being represented as a vector, through the matrix (i.e., to maximize $\vec{s}^T M_r \vec{o}$ ). Most following KGE models, such as TransR, DistMult, and ComplEx, were further developed on top of these two basic models.

One limitation of applying KGE methods on geospatial knowledge graphs is that they can hardly utilize the spatial and temporal information for geospatial prediction, as they are mostly designed for non-spatial or general tasks. To address this gap, researchers have introduced spatially explicit KGE, a sub-field of spatially explicit machine leaning. In this approach, geospatial knowledge from the nodes or edges of the graph is explicitly considered to enhance the performance of downstream tasks. An illustration of this is TransGeo, a spatially explicit variant of TransE designed for geospatial question answering (Mai et al., 2020). TransGeo incorporates a distance-weighting scheme to explicitly integrate the distance decay effect into the conventional TransE model, resulting in substantial improvements across various geospatial tasks such as link prediction and query relaxation.

Despite significant advancements in geospatial knowledge graphs in recent years, this field is still in its early stages. There exist challenges, but also opportunities, that must be addressed to fully realize the potential of this promising technology in geography and beyond. Firstly, there remains a gap in fundamental research that connects theories in geospatial semantics with the technical capabilities of knowledge graphs. Concepts such as semantic reference systems, image schemata, and cognitive space have been proposed for modeling geographic information, particularly in human geography and social science. However, their implementations in GIS systems have not yet been realized, and geospatial knowledge graphs could potentially serve as a new paradigm to bridge this gap. Secondly, while GeoSPARQL standardizes the representation of geospatial information in a knowledge graph, it is limited in consolidating multimodal geospatial information, including remotely sensed images, human mobilities, vague regions, and unstructured geo-text. Lastly, there is a need to create portable and accessible models and tools, similar to spatial statistical modeling packages in R and Python. These resources should empower domain scientists to easily access data from geospatial knowledge graphs and harness the latest advancements in spatially explicit KGE to uncover valuable insights from data.

**SEE ALSO:**
Spatial Data Science
Data Structure: Spatial Data on the Web
GeoAI and Deep Learning
Geospatial Semantic Web
Interoperability of Representations
Machine Learning
Ontology: Theoretical Perspectives
Open Geospatial Consortium standards
Platial
Qualitative Spatial and Temporal Representation and Reasoning

## Further Readings

## Key Words

Spatial databases, spatiotemporal data, representation, geographic technologies, data acquisition, GIScience, knowledge